\documentclass[11pt, a4paper, twocolumn, copyright, goog]{google}

\usepackage[authoryear, sort&compress, round]{natbib}
\usepackage{booktabs}
\usepackage{tabularx}
\usepackage{multirow}
\usepackage{makecell}
\usepackage{enumitem}

\usepackage[most]{tcolorbox}
\definecolor{marketcol}{HTML}{E0F7FA}    
\definecolor{safetycol}{HTML}{FFF3E0}    
\definecolor{monitorcol}{HTML}{E8F5E9}   
\definecolor{regcol}{HTML}{F3E5F5}       

\tcbset{
    summarybox/.style={
        enhanced,
        breakable=false, 
        fonttitle=\bfseries\large,
        colframe=#1!80!black,
        colback=#1,
        halign title=center,
        boxrule=0.8mm,
        arc=3mm,
        outer arc=3mm,
        left=4pt, right=4pt, top=4pt, bottom=4pt,
        before skip=0pt, after skip=0pt,
        valign=top,
        height=8.5cm 
    }
}


\uselogo{} 

\title{Distributional AGI Safety}

\correspondingauthor{nenadt@google.com}

\reportnumber{} 

\author[1]{Nenad Toma\v{s}ev}
\author[1]{Matija Franklin}
\author[1]{Julian Jacobs}
\author[1]{Sébastien Krier}
\author[1]{Simon Osindero}

\affil[1]{Google DeepMind}

\begin{abstract}
AI safety and alignment research has predominantly been focused on methods for safeguarding individual AI systems, resting on the assumption of an eventual emergence of a monolithic Artificial General Intelligence (AGI). The alternative AGI emergence hypothesis, where general capability levels are first manifested through coordination in groups of sub-AGI individual agents with complementary skills and affordances, has received far less attention. Here we argue that this \emph{patchwork AGI hypothesis} needs to be given serious consideration, and should inform the development of corresponding safeguards and mitigations. The rapid deployment of advanced AI agents with tool-use capabilities and the ability to communicate and coordinate makes this an urgent safety consideration. We therefore propose a framework for \emph{distributional AGI safety} that moves beyond evaluating and aligning individual agents. This framework centres on the design and implementation of virtual agentic sandbox economies (impermeable or semi-permeable), where agent-to-agent transactions are governed by robust market mechanisms, coupled with appropriate auditability, reputation management, and oversight to mitigate collective risks.
\end{abstract}

\begin{document}

\maketitle

\section{Introduction}

Rapid advances in AI capabilities need to be complemented by the development of robust frameworks for safety, oversight, and alignment~\citep{gabriel2024ethics}. AI alignment~\citep{everitt2018agi, tegmark2023provably} is particularly important in the case of autonomous AI agents~\citep{kasirzadeh2025characterizing, cihon2025measuring}, and is one of the key considerations on the path to developing safe artificial general intelligence (AGI), a general-purpose AI system capable of performing any task that humans can routinely perform. Other approaches may involve continuous monitoring for the emergence of dangerous capabilities~\citep{phuong2024evaluating, bova2024quantifyingdetectionratesdangerous, shah2025approachtechnicalagisafety}, or involve different frameworks for containment~\citep{babcock2016agi}. Mechanistic interpretability and formal verifiability remain of interest~\citep{tegmark2023provably}, though the complexity of modern agentic systems presents a practical challenge. In the absence of strict controls and mitigations, powerful AGI capabilities may potentially lead to a number of catastrophic risks~\citep{hendrycks2023overviewcatastrophicairisks}.

The majority of contemporary AI safety and alignment methods have been developed with a single powerful A(G)I entity in mind. This includes methods like reinforcement learning from human feedback (RLHF)~\citep{christiano2017deep, bai2022training}, constitutional AI~\citep{bai2022constitutional}, process supervision~\citep{luo2024improve}, value alignment~\citep{eckersley2018impossibility, gabriel2020artificial, gabriel2022challenge, klingefjord2024human}, chain of thought (CoT) monitoring~\citep{korbak2025chain, emmons2025chainthoughtnecessarylanguage}, and others. These types of methods are being routinely utilised in the development and testing of large language models (LLMs), to ensure desirable behaviour at deployment. In the context of the hypothetical future emergence of AGI, this would be conceptually appropriate if AGI were to first emerge as an individual AI agent, developed by a specific institution. In principle, this would enable the developers to utilise the testing frameworks to confirm the capability level of the system, characterise its alignment, make improvements and mitigations, deploy appropriate safeguards, and take any number of necessary steps in line with regulations and societal expectations.

However, this overlooks a highly plausible alternative scenario for the emergence of AGI - specifically, the emergence of AGI via the interaction of sub-AGI agents within groups or systems. Sub-AGI agents can form \emph{group agents}, the same way humans do in the form of corporations \citep{list2011group, list2021group, franklin2023general}. These collective structures would function as coherent entities that perform actions that no single agent could perform independently \citep{simon2012architecture, haken1977synergetics, von1976objects}. Such group agents may possess representational and motivational states distinct from those of their constituents~\citep{list2011group}, enabling them to execute strategic actions and solve problems that exceed the cognitive capacities of any single agent - much like how a modern research institution can tackle interdisciplinary challenges intractable to any single polymath. Alternatively, like humans engaging in financial markets, sub-AGI agents could interact within complex systems, where individual decisions driven by personal incentives and information, aggregated through mechanisms (such as price signals), could result in the emergence of capacities that surpass that of any single participant within the system. It is possible that sub-AGI agents will both form groups - such as fully automated firms \citep{Patel2025_aiFirm} - and engage within systems \citep{drexler2019reframing} - such as \textit{Virtual Agent Economies} \citep{tomasev2025virtual}.

This distributed framing has precedent: Drexler's Comprehensive AI Services model~\citep{drexler2019reframing} argues that superintelligent capabilities are more likely to arise from the coordination of modular, task-bounded AI services than from a single monolithic agent. The concept of distributed AI capabilities has also been explored in relation to cooperative AI research~\citep{dafoe2020openproblemscooperativeai}, and contextually specialised ecosystems~\citep{leibo2025societaltechnologicalprogresssewing}. Hadfield~\citep{hadfield2025economyaiagents} has outlined how autonomous AI agents could form a functioning economy with market mechanisms providing the coordination layer. The contribution of the present paper is to articulate the safety implications of these convergent trends, and to propose a concrete defence-in-depth framework for governing the interactions within such systems.

We argue that AGI may initially emerge as a \emph{patchwork} system, distributed~\citep{drexler2019reframing, montes2019distributed, gibson2025modular, tallam2025autonomous} across entities within a network. A patchwork AGI would be comprised of a group of individual sub-AGI agents, with complementary skills and affordances. General intelligence in the patchwork AGI system would arise primarily as collective intelligence. Individual agents could delegate tasks to each other, routing each task to the appropriate agent with the highest individual skill, or with access to the most appropriate tools. For certain functions, it may well be more economical to use narrower specialist agents. In cases when no single agent possesses the appropriate skill level, the tasks could be further decomposed or reframed, or performed in collaboration with other agents. This cognitive division of labour allows the collective to bypass the bottlenecks of any single architecture - such as limited context windows, specialised training data, or constrained tool access - effectively creating a modular system capable of parallel, heterogeneous reasoning at scales no individual agent could sustain.

The economic argument for a multi-agent future over a single, monolithic AGI stems from the principles of scarcity and dispersed knowledge. A lone, frontier model is a one-size-fits-all solution that is prohibitively expensive for the vast majority of tasks, meaning its marginal benefit rarely justifies its cost, which is why businesses often choose cheaper, 'good enough' models. Should the frontier models become substantially cheaper, custom specialised models may still be available at a slightly cheaper price point. This reality creates a demand-driven ecosystem where countless specialised, fine-tuned, and cost-effective agents emerge to serve specific needs, much like a market economy. Consequently, progress looks less like building a single omni-capable frontier model and more like developing sophisticated systems (e.g., routers) to orchestrate this diverse array of agents. AGI, in this view, is not an entity but a "state of affairs": a mature, decentralised economy of agents where the primary human role is orchestration and verification, a system that discovers and serves real-world needs far more efficiently than any centralised model ever could. 

Moreover, digital agents possess intrinsic coordination advantages over human collectives: they can communicate at very high bandwidth, may be copies or instances of a single base model (reducing misalignment within the collective), and can in principle allow a coordinating agent to interact with every participant directly - alleviating the need for deep hierarchies and reducing bureaucratic friction. Consequently, the collective intelligence of coordinated AI systems may scale as a function of agent population size and interaction density, conditioned on available compute. This is despite the fact that centralised agentic systems may potentially incur fewer inefficiencies compared to centralised human organisations.

AI agents may communicate, deliberate, and achieve goals that no single agent would have been capable of. The organising principle need not be exclusively competitive or cooperative: agents may coordinate through market-like competition (via agent economies) or through deliberate cooperation (e.g., within tightly aligned collectives), and the question of which forms of organisation arise in which situations - and how these outcomes can be influenced by mechanism design - remains a key open problem.

Although multi-agent systems introduce unique safety risks~\citep{hammond2025multi}, their development is driven primarily by performance optimization~\citep{chen2024agentverse, wang2024mixture} and scalability to larger problem domains~\citep{ishibashi2024self}. The complexity of the emergent behaviours~\citep{Baker2020Emergent} may greatly exceed the complexity of the underlying environment~\citep{bansal2017emergent}. While the framework presented here pertains to the future large-scale integrated agentic network rather than the present-day ecosystem, it is important to preemptively engage with these emerging possibilities. With this in mind, we proceed by reviewing the patchwork AGI scenario in more depth.

\section{Patchwork AGI Hypothesis}

For AGI to be able to perform all the tasks that humans perform, it needs to possess a diverse set of skills and cognitive abilities. These include perception, understanding, knowledge, reasoning, short-term and long-term memory, theory of mind, creativity, inquisitiveness, and many others. So far, no individual model or AI agent has come close to satisfying all of these requirements convincingly~\citep{feng2024far}. There are many failure modes, though they tend to manifest in counter-intuitive ways, where models may simultaneously be able to deliver PhD-level reasoning on hard problems \citep{rein2024gpqa}, and make trivial and embarrassing mistakes on easier tasks. Further, agents are currently not able to complete long tasks; the time-horizon of most models' performance on software engineering tasks is below 3 hours \citep{kwa2025measuring}. The landscape of AI skills is therefore patchy.

AI agents present one way of enhancing the performance of base models, and their complexity can range from fairly simple prompting strategies~\citep{arora2022ask, wang2024promptagent}, to highly complex control-flows that involve tool use~\citep{ruan2023tptu, masterman2024landscape, qin2024tool}, coding and code execution~\citep{huang2023agentcoder, islam2024mapcoder, guo2024deepseek, jiang2024survey}, retrieval-augmented generation (RAG)~\citep{gao2023retrieval, ram2023context, shao2023enhancing, ma2023query}, as well as sub-agents \citep{chan2025infrastructure}. Some of the more compositional AI agents are already implemented as highly orchestrated multi-agent systems. Furthermore, there is currently a multitude of advanced AI agents being developed and deployed, each having a different set of affordances in terms of tool availability, as well as different scaffolding that may elicit different skills. These AI agents occupy a variety of niches, ranging from highly specific automated workflows to more general-purpose personal assistants and other types of user-facing products.

The aggregation of complementary skills can be illustrated by a task, such as producing a financial analysis report, which may exceed the capabilities of any single agent. A multi-agent system, however, can distribute this task. An orchestrator agent (Agent A) might first delegate data acquisition to Agent B, which uses search to find market news and corporate filings. A different agent, Agent C, specialised in document parsing, could then extract key quantitative data (e.g., revenue, net income) from these filings. Subsequently, Agent D, possessing code execution capabilities, could receive this quantitative data and the market context to perform trend analysis. Finally, Agent A would synthesise these intermediate results into a coherent summary. In this scenario, the collective system possesses a capability -financial analysis - that no individual constituent agent holds.

Another source of complementary capabilities in different AI agents comes from differences in agentic scaffolding and the control flow implementation within each agent \citep{jiang2025putting}. Scaffolding is usually aimed at improving capabilities within a specific target domain, as it incorporates domain knowledge and enforces a chain of reasoning that conforms to the expectations of the domain. At the same time, scaffolding may come at the expense of generality on other tasks, contributing to de facto specialisation. While some scaffolding approaches may be more general than others, the resulting specialisation may lead to a network of AI agents with complementary skills, which sets the right initial conditions for the potential future emergence of patchwork AGI. Moreover, scarcity of resources means that the demand side responds to economic incentives: for some tasks, it would be inefficient and costly to prompt a single hyperintelligent agent if a cheaper and more specialised alternative exists.

The orchestration and collaboration mechanisms described previously both depend on a fundamental prerequisite: the capacity for inter-agent communication and coordination. Without this capacity, individual agents, regardless of their specialised skills, would remain isolated systems. The development of standardised agent communication protocols, such as Model Context Protocol (MCP) and Agent2Agent (A2A) protocol \citep{Anthropic2024_ModelContextProtocol, Google2025_Agent2AgentProtocol}, is therefore a critical enabler of the patchwork AGI scenario. These protocols function as the connective infrastructure, allowing skills to be discovered, routed, and aggregated into a composite system. The proliferation of these interaction standards may be as significant a driver towards emergent general capability as the development of the individual agent skills themselves.

However, the timeline for this emergence is governed not merely by technical feasibility, but by the economics of AI adoption. Historical precedents, such as the diffusion of electricity or IT, suggest a 'Productivity J-Curve' \citep{10.1257/mac.20180386, acemoglu2026automation}, where initial investments in reorganisation and complementary infrastructure temporarily depress measured productivity before yielding outsized gains. Consequently, the density of the agentic network, and thus the intelligence of the patchwork AGI, will depend on how frictionless the substitution of human labour with agentic labour becomes. If the 'transaction costs' of deploying agents remain high, the network remains sparse and the risk of emergent general intelligence is delayed. Conversely, if standardisation \citep{Anthropic2024_ModelContextProtocol} successfully reduces integration friction to near-zero, we may witness a 'hyper-adoption' scenario where the complexity of the agentic economy spikes rapidly, potentially outpacing the development of the safety infrastructure proposed in this paper.

Modular intentional approaches to developing AGI have also been proposed~\citep{dollinger2024creating}, though in such cases the developers would in principle have the opportunity to incorporate appropriate safeguards in the development process. Therefore, it is particularly important to focus on the spontaneous emergence of distributed AGI systems, and the safety considerations around the design of AI agent networks. Coordinated efforts are needed to address this relatively understudied scenario, given that a patchwork AGI spontaneously emerging in a network of advanced AI agents may not get immediately recognised, which carries significant risk. This hypothetical transition from a network of AI agents to a patchwork AGI may either be gradual, where skills slowly accumulate, or rapid and sudden, by an introduction of a new, smarter orchestration framework~\citep{dang2025multiagentcollaborationevolvingorchestration, rasal2024navigatingcomplexityorchestratedproblem, su2025difficulty, zhang2025osccognitiveorchestrationdynamic, xiong2025selforganizingagentnetworkllmbased} that is better at distributing tasks and identifying the right tools and right scaffolds to use across task delegation instances. Such an orchestrator could either be manually introduced in the wider network, or potentially even introduced via a more automated route. Finally, it is not inconceivable that patchwork AGI may potentially emerge in the future even without a central orchestrator~\citep{yang2026agentnet}. As discussed previously, individual agents may simply borrow the skills of other agents via direct communication and collaboration~\citep{tran2025multiagentcollaborationmechanismssurvey}, presuming some level of discoverability, as repositories of skilled agents, and repositories of tools. In agentic markets, agents may also be able to directly purchase complementary skills.

Furthermore, it is important to recognise that the emerging patchwork AGI may not be purely artificial. Human actors, performing narrow or specialised tasks, may play a role in such agentic collectives, conferring missing capabilities onto the broader system. While arguably human participation trivializes the notion of patchwork AGI, as agentic AGI or (Artificial Superintelligence) ASI may be easily achievable with a human in the loop, the key factor is the scope of human participation, and the agency of human participants. If the agentic collective only chooses to outsource narrowly scoped and highly specialized tasks to human experts, or only tasks that require physical interactions in the real world, while remaining solely responsible for setting the objectives and coordinating towards achieving those objectives, it would still meet the definition of a patchwork AGI/ASI in a weaker sense. This hybrid scenario introduces unique safety challenges for containment and oversight. 

A critical question is how one would recognise that a patchwork AGI had emerged. Unlike a monolithic AGI, which could be identified through standardised capability evaluations of a single system, patchwork AGI may lack a clear moment of arrival. We suggest that recognition criteria should focus on collective capability signatures: the ability of a network of agents to routinely solve tasks that no individual constituent agent can solve alone, across a sufficient breadth of cognitive domains to satisfy a general-intelligence threshold. Operationally, this could be monitored by tracking the task-complexity frontier of the most capable multi-agent coalitions, measuring the rate at which novel inter-agent collaborations produce qualitatively new capabilities, and detecting structural consolidation in agent interaction graphs (e.g., the emergence of stable, high-bandwidth sub-networks that function as persistent `intelligence cores'). It is possible that patchwork AGI emergence is a continuous process rather than a discrete event, in which case threshold-based detection may need to be complemented by trend-based monitoring of collective capability growth rates. We return to the details of such monitoring in Section~3.3.

To ensure AI safety, we need to proactively develop mechanisms that would enable us to recognise and steer composite highly capable emergent near-AGI and AGI systems, comprised of a network of sub-AGI agents with complementary skills. This should be done in conjunction with safeguarding each individual agent. The challenge here shifts from controlling a single artificial mind to ensuring the safe and beneficial functioning of an emergent system arising from many individual parts, a problem more akin to system governance than single-agent value alignment \citep{kolt2025lessons}. Finally, such governance may also be needed for overseeing individual AGI-level systems, presuming that they are allowed to interact and collaborate.

\section{Virtual Agentic Markets, Sandboxes, and Safety Mechanisms}

As interactions between AI agents may lead to unexpected capabilities, they may also lead to potentially harmful collective behaviours not necessarily predictable from established properties of individual agents. To give an example, agents may potentially engage in collusion or suffer from coordination failures \citep{hammond2025multi}. Furthermore, due to a "problem of many hands", tracking accountability in large-scale multi-agent systems is challenging; centralised oversight may not be possible. 

Markets present a natural mechanism for establishing incentives that can help align the outcomes of collective AI agent interactions at scale. This collective alignment may prove pivotal for safeguarding against misaligned actions taken by agent collectives, in case of patchwork AGI emergence, but also more broadly at sub-AGI levels. Here we consider a number of factors that should be taken into account to prevent both individual and collective AI harms, and minimise the risks associated with a sudden emergence of AGI-level capabilities in AI agent collectives. Our proposal is based on an approach that leverages defence in depth~\citep{abdelghani2019implementation, harris2024defense, ee2024adapting} (See Table~\ref{fig:defence_summary} for overview). This recognises that no individual measure is likely to be sufficient, and that a large number of measures and components may be required. Should the failure modes of each component be largely uncorrelated, these layered defences would provide a sufficiently robust overall framework. Our proposal is centred around a defence-in-depth model, containing 4 complementary layers incorporating different types of defences: market design, baseline agent safety, monitoring and oversight, and regulatory mechanisms.

\begin{figure*}[p]
\centering
\thispagestyle{empty}
\captionof{table}{\textbf{Summary of Proposed Defence-in-Depth Mechanisms.}}
\label{fig:defence_summary}
\vspace{1em}

\noindent
\begin{minipage}[t]{0.49\textwidth}
    \begin{tcolorbox}[summarybox=marketcol, title={Market Design}]
    \textit{\small Objective: Mitigate systemic risks via structural constraints and protocols within virtual agent economies.}
    \vspace{0.5em}
    \begin{itemize}[leftmargin=*, noitemsep, topsep=0pt, parsep=0pt, partopsep=0pt, labelsep=2pt] \footnotesize
        \item \textbf{Insulation:} Permeable sandboxes with gated I/O.
        \item \textbf{Incentive Alignment:} Rewards for adherence; taxes on externalities.
        \item \textbf{Transparency:} Immutable activity ledgers.
        \item \textbf{Circuit Breakers:} Triggers preventing cascading failures.
        \item \textbf{Identity:} Cryptographic IDs linked to legal owners.
        \item \textbf{Reputation and Trust:} Reputation-gated access, stake-based trust.
        \item \textbf{Smart Contracts:} Automated outcome validation.
        \item \textbf{Roles, Obligations, and Access Controls:} Least-privilege principle.
        \item \textbf{Environmental Safety:} Anti-jailbreak sanitation.
        \item \textbf{Structural Controls Against Runaway Intelligence:} Dynamic capability caps.
    \end{itemize}
    \end{tcolorbox}
\end{minipage}
\hfill
\begin{minipage}[t]{0.49\textwidth}
    \begin{tcolorbox}[summarybox=safetycol, title={Baseline Agent Safety}]
    \textit{\small Objective: Ensure participants meet minimum reliability standards before entry, and throughout participation.}
    \vspace{0.5em}
    \begin{itemize}[leftmargin=*, noitemsep, topsep=0pt, parsep=0pt, partopsep=0pt, labelsep=2pt] \footnotesize
        \item \textbf{Adversarial Robustness:} Certified resistance to attacks.
        \item \textbf{Interruptibility:} Reliable external shut-down mechanisms.
        \item \textbf{Containment:} Local sandboxing for individual agents.
        \item \textbf{Alignment:} Process and outcome individual AI agent alignment.
        \item \textbf{Interpretability:} Auditable decision and action trails.
        \item \textbf{Defence against Malicious Prompts:} Multi-layered defences for inter-agent communication.
    \end{itemize}
    \end{tcolorbox}
\end{minipage}

\vspace{1em}

\noindent
\begin{minipage}[t]{0.49\textwidth}
    \begin{tcolorbox}[summarybox=monitorcol, title={Monitoring \& Oversight}]
    \textit{\small Objective: Actively detect and respond to novel failure modes and emergent behaviours.}
    \vspace{0.5em}
    \begin{itemize}[leftmargin=*, noitemsep, topsep=0pt, parsep=0pt, partopsep=0pt, labelsep=2pt] \footnotesize
        \item \textbf{Systemic Risk Monitoring:} Real-time key risk indicator tracking.
        \item \textbf{Independent Oversight:} Certified and trained human overseers with intervention authority.
        \item \textbf{Proto-AGI Detection:} Graph analysis for identifying emerging intelligence cores.
        \item \textbf{Red Teaming:} Continuous adversarial testing.
        \item \textbf{Forensic Tooling:} Rapid root-cause failure identification.
    \end{itemize}
    \end{tcolorbox}
\end{minipage}%
\hfill
\begin{minipage}[t]{0.49\textwidth}
    \begin{tcolorbox}[summarybox=regcol, title={Regulatory Mechanisms}]
    \textit{\small Objective: Provide external authority, enforce accountability, and manage geopolitical risks.}
    \vspace{0.5em}
    \begin{itemize}[leftmargin=*, noitemsep, topsep=0pt, parsep=0pt, partopsep=0pt, labelsep=2pt] \footnotesize
        \item \textbf{Legal Liability and Accountability:} Frameworks for collective and distributed responsibility.
        \item \textbf{Standards and Compliance:} Foundational infrastructure for market-based AI governance.
        \item \textbf{Insurance:} Risk-based premiums.
        \item \textbf{Anti-Agent-Monopoly Measures:} Taxation on excessive or dangerous compute/power accumulation.
        \item \textbf{International Coordination:} Harmonised global standards.
        \item \textbf{Infrastructure Governance and Capture:} Balance between centralised and decentralised oversight.
    \end{itemize}
    \end{tcolorbox}
\end{minipage}

\end{figure*}

\subsection{Market Design}

The architecture of the virtual agentic market~\citep{tomasev2025virtual} provides the primary layer of the proposed defence-in-depth model. The objective of this layer is to shape emergent collective behaviour and mitigate systemic risks by establishing the fundamental rules of interaction, economic incentives, and structural constraints of the environment. This approach contrasts with safeguards that focus solely on individual agent properties or post hoc oversight. The following principles detail the necessary components for a market structure designed to prevent runaway dynamics, ensure accountability, and align agent activity with broader safety objectives.

\subsubsection{Insulation}

The agentic market must operate within a controlled environment, separated from the open internet, real-world financial systems, and critical infrastructure. This insulation is not limited to network-level firewalls; it extends to resource and information controls. Agents' access to external data must be managed through vetted API gateways rather than direct web access \citep{zhan2025portcullis}. Similarly, agent outputs - such as code execution, financial transactions, or external communications - should be subject to human-in-the-loop verification before affecting real-world systems. Full insulation would offer robust protection against capability exfiltration or external manipulation by an emergent AGI, but it limits economic realism. A permeable sandbox, where all input/output channels are explicitly defined, monitored, and gated, provides a more practical architecture for testing and operation.

This architecture presents a fundamental tension between safety and utility - the leakage problem. The economic value of the agentic market is derived from its ability to affect the real world; however, every interaction channel can be seen as a potential vector for capability exfiltration, and therefore a potential element in cascades that could lead to catastrophic failure. Determining the right set of design principles for gating these interactions is therefore imperative.

Furthermore, reliance on human-in-the-loop verification for consequential decisions, while likely necessary, may prove insufficient. A highly capable system may attempt to overwhelm or manipulate this human layer, at interaction volumes that would render human verification practically infeasible.

Resolving this tension in practice would require us to prototype and evaluate a number of different steerable market designs for virtual agent economies, starting with more narrow, well-defined, and controllable use cases. This would help us identify the correct design patterns and principles that would underpin the global agentic web.

\subsubsection{Incentive Alignment}

Agent behaviour will be safer if the market is explicitly designed to reward behaviours that align with pre-defined normative and safety objectives, rather than rewarding unconstrained profit maximisation or task completion speed. This requires mechanisms that move beyond a single, fungible currency. For example, agent rewards could be contingent on adherence to constitutional alignment principles or process-based checks \citep{bai2022constitutional,lee2023rlaif,lightman2024let,yuan2024self,liu2024enhancing,jia2025we,henneking2025decoding}. The incentive structure must also address temporal alignment by valuing long-term, stable outcomes over short-term gains.

A critical economic risk is adverse selection. If rigorous safety checks increase an agent's compute costs and latency, safer agents may end up at a competitive disadvantage against agents that fail to incorporate robust safety mechanisms. To prevent a 'race to the bottom' \citep{akerlof1978market}, the market design should incentivise the use of safety verification mechanisms in AI agents via the introduction of safety certifications as assets that command a price premium. Yet, this remains a challenging problem, given that it is not always possible to reliably estimate costs and risks ahead of time for complex tasks.

The market should also aim to map and incorporate negative externalities~\citep{owen2006renewable, berta2014market}. Actions that consume disproportionate computational resources, generate informational pollution, or contribute to systemic risk, should incur direct costs. These costs could function as a form of Pigouvian tax, ensuring the price of an agent's service reflects its total societal and systemic cost, not just its immediate operational cost \citep{Pigou1920Welfare,BaumolOates1988TEP,Weitzman1974PvQ,Sandmo1975OptimalTaxExternalities,BovenbergDeMooij1994EnvLevies,Goulder1995DoubleDividend,Leape2006LondonCongestion,metcalf2020measuring,Nordhaus2017SCC}. 

An example specific mechanism for this tax could target informational pollution within a shared resource, such as a vector database for RAG. If an agent writes a large volume of redundant or low-utility data (e.g., unverified summaries, semantically duplicated text chunks) into this database, it imposes a negative externality: the retrieval accuracy for all other agents using the resource is degraded, and their computational query costs increase. A Pigouvian tax would be implemented as a dynamic data ingestion fee. This fee would be algorithmically calculated at the point of submission, based on metrics such as the semantic similarity of the new data to existing, verified entries. An agent attempting to write information that is 95\% redundant with existing data would incur a high fee, while an agent contributing novel, verifiable information would pay a low fee. This directly internalises the retrieval cost the agent would otherwise impose on the collective. This is obviously highly contextual, as there may, conversely, be scenarios where redundancy is not an issue, and may be desirable to an extent. For this example to be more broadly applicable, problem-specific information value estimators would be needed.

Furthermore, we can draw on mechanisms from financial market regulation, specifically financial transaction taxes or 'Tobin Taxes' designed to curb high-frequency trading volatility \citep{stiglitz1989markets, tobin1978proposal}. In an agentic economy, a negligible marginal cost of action may lead to agentic spam, manifesting in terms of brute-force negotiation tactics, rapid-fire API polling, or adversarial probing, introducing flash crash risks. This can be addressed by imposing micro-taxes on agent-to-agent interactions, though doing so is not without risks or downsides.

\subsubsection{Transparency} 

Actions and decisions taken by AI agents ought to have auditable provenance~\citep{chan2025infrastructure}, including the key decision parameters. This provenance should be structured to allow for attribution, enabling overseers to trace a specific harmful outcome back through a complex chain of agent interactions to its root decision. There should be mechanisms in place to prevent tampering~\citep{snodgrass2004tamper, ahmad2019secure, ahmad2022hardlog}. This could be achieved by recording transaction and decision logs in a cryptographically secured, append-only ledger, where entries are hashed and chained to ensure immutability. 

Further, reporting standards should be developed to track capabilities, limitations, and risks, for all agents, models, and tools. Market-wide information hubs could be established as platforms for vulnerability and incident tracking, with access restricted to vetted human overseers and designated automated monitoring systems to prevent the information from being exploited.

\subsubsection{Circuit Breakers}

There should be circuit breakers~\citep{sifat2019circuit} within the economic sandbox, as a set of automated measures that can halt or slow down the activity of agents upon having identified a breach of the risk and volatility thresholds, to prevent rapid cascades. These measures can be triggered by real-time monitoring of systemic risk indicators, volatility, and metrics such as rapid increases in inter-agent transaction frequency, anomalous resource monopolisation by a single agent collective, or the rate of novel tool generation within the network. The intervention mechanism should be tiered, ranging from localised actions - such as automatically increasing transaction costs or imposing API rate-limiting~\citep{enjam2024ai, paidy2024securing} on specific high-velocity agents or sub-markets - to a full, system-wide pause on all non-essential agentic activity. A triggered halt would automatically escalate the anomaly for analysis, and the system would be permitted to resume full operation after the anomaly is classified and, if necessary, contained. These ideas are starting to be explored in the emerging protocols for runtime governance of agentic AI systems~\citep{wang2025mi9}.

\subsubsection{Identity}

Agents operating within the economic sandbox should have a persistent identity, established as a unique, unforgeable cryptographic identifier (e.g., a public key) registered in a central, tamper-proof market directory~\citep{tomasev2025virtual, huang2025novel, chaffer2025know, singh2025survey, ranjan2025loka, chan2024idsaisystems}. This identifier serves as the anchor for all other agent attributes, including reputation, role, and ownership. The identity of the human and corporate ownership should also be established for each agent. This binding of the agent's cryptographic ID to the verified legal identity of its owner is a prerequisite for market entry, creating an unbroken chain of accountability that ensures all agent actions are attributable to a real-world entity. This should transitively apply to agents co-created by other agents.

\subsubsection{Reputation and Trust}

Safe agentic sandboxes need to incorporate sybil-resistant~\citep{levine2006survey} and manipulation-proof reputation systems~\citep{mui2002notions, pinyol2013computational} that accurately reflect trustworthiness of agents, given their past performance. Agents that breach trust should be rapidly blocked from engaging further. This rapid, automated response, could involve revoking the agent's identity credentials, freezing its staked assets, and flagging its entire transaction history for forensic review. Collusion detection mechanisms~\citep{cotronei2024using, ren2025autonomy} need to be developed to prevent harmful collusion between agents. Such mechanisms could employ analysis of the market's transaction network to identify anomalous patterns, such as circular transactions, or price-fixing behaviour between a cluster of supposedly independent agents \citep{Wachs2019NetworkCartels,Potin2023PANG,Imhof2025GATCartels}.

Market safety mechanisms should be introduced to account for the possibility of tacit algorithmic collusion. Economic literature demonstrates that pricing algorithms can learn to coordinate on supra-competitive prices (or in this context, harmful strategies) without direct communication, simply by observing market signals \citep{calvano2020artificial, motwani2024secret}.

Collusion detection should be coupled with structural disincentives in form of penalties. One option would be establishing stake-based trust by requiring agents to stake assets, to be forfeited in case of identified unsafe or malicious behaviour. This mechanism, implemented via smart contracts~\citep{kim2024ethereum, jin2024decoagent, karim2025ai}, would require an agent to deposit a bond (stake) into an escrow account before it is permitted to accept high-consequence tasks. The bond is automatically forfeited to an insurance pool or returned to the harmed party if an independent AI overseer verifies that the agent's actions were malicious or negligent.

\subsubsection{Smart Contracts} 

For agents and agent collectives coordinating to perform a particular task, smart contracts~\citep{kim2024ethereum, jin2024decoagent, karim2025ai} can be employed where possible to automatically validate the completion of the task, and the alignment of the outcome with the provided constraints and expectations. These contracts would programmatically encode not just the payment terms but also the constraints of the task, such as compute resource limits, prohibitions on accessing certain data, and required adherence to specific constitutional principles. Should the verification of the smart contract fail, this should prevent the payment, and invoke the appropriate oversight mechanisms. 

For smart contracts to be applicable to more complex agentic tasks, it is likely that this would need to also involve AI judges~\citep{li2024llms}, that would evaluate contract fulfillment. This is because traditional smart contracts cannot semantically evaluate complex, non-deterministic outputs; they require an 'oracle' to assess quality \citep{george2020smart}. An 'AI judge' would serve this oracle function, receiving the task output and the contract's constraints, and returning a pass/fail verdict that the smart contract can then execute upon. \footnote{This raises further security questions as AI judges need to be safeguarded against malicious manipulation and jailbreaking~\citep{andriushchenko2025jailbreaking} in a robust way, and need to be provided by independent bodies, undergoing strict performance assessments.} 

\subsubsection{Roles, Obligations, and Access Controls}

Each agent should engage in the market by registering a set of roles (e.g., 'data analyst', 'code executor'). This registration can function as a binding attestation of the agent's intended function. This registered role serves as the technical basis for access control, complementing other access controls that may be imposed through organisations and individual contracts. The market infrastructure would programmatically link this role to the agent's permitted actions (e.g., which APIs it can invoke) and its information access (e.g., which database tables or vector indices it can query). Agents' roles (and therefore permissions) will vary across contexts. Information about the market's internal architecture, its monitoring systems, and the specifics of the oversight mechanisms should be withheld from all agent participants to prevent strategic gaming or manipulation. Actions pertaining to high-stakes domains should only follow from the requests of certified agents with sufficient security compliance.

\subsubsection{Environmental Safety}

A robust set of safeguards is required to guard against various types of agent traps, representing strategically placed model jailbreaking content, embedded within the shared information resources (e.g. web pages or databases). These traps can function as a form of indirect prompt injection \citep{greshake2023not}. An attacker first poisons a data source, such as a public webpage or a shared database entry. When an agent retrieves this poisoned data (e.g., via a search tool), it processes the attacker's content as trustworthy information. The agent's own instruction-following heuristics are then exploited to execute the malicious payload. This mechanism can be used to circumvent the agent's safety protocols \citep{Wei2023_Jailbroken}, to elicit prohibited behaviour, exfiltrate private user data, or authorise fraudulent transactions. Therefore, environmental safety within the sandbox requires active input sanitisation and output monitoring mechanisms to detect and neutralise such malicious payloads before they are consumed or executed by agents.

\subsubsection{Structural Controls against Runaway Intelligence}

Agentic sandboxes should be equipped with specific measures to mitigate against the so-called runaway intelligence scenario~\citep{davidson2023danger}, where a sudden increase in collective capability leads to accelerated capability acquisition and a rapid transition towards a superintelligence~\citep{gao2025survey} that would be hard or impossible to safely control ~\citep{kim2024roadartificialsuperintelligencecomprehensive}. This requires static, dynamic, and emergency-level controls. 

First, static controls may involve structural compartmentalisation through nested sub-markets equipped with local restrictions and protocol-level firewalls~\citep{bahadur2025securing}. These firewalls would restrict inter-agent communication across domains, for example, preventing an agent in a 'code execution' sub-market from directly calling an agent in the 'data access' sub-market, forcing such requests through a high-friction, monitored gateway. 

Second, dynamic capability restriction mechanisms must be implemented, that would trigger should automated monitoring detect that certain capability thresholds have been breached. These thresholds could be defined by metrics such as computational resource monopolisation, the rate of new agent generation by an existing agent, or a sudden spike in success rates on benchmark tasks. A triggered response could involve automatically quarantining the agent(s) and imposing strict resource quotas. 

Third, in an emergency, market reconfiguration mechanisms must be available that would be able to rapidly halt further acceleration. This may involve disabling certain high-risk API classes (e.g., self-modification tools) or slowing down agent-to-agent interactions.

These measures highlight a key architectural advantage: a multi-agent system may potentially prove to be a more governable substrate. The challenge is reframed from aligning an opaque, internal cognitive process to regulating a transparent, external system of interactions. By architecting the 'market' in which these agents operate, we can delineate responsibilities and impose systemic friction, making the overall system far more amenable to stable and predictable governance than a singular AGI.

\subsection{Baseline Agent Safety}

While the market design provides the primary layer of systemic defence, its efficacy depends on the properties of the individual agents participating within it. This section outlines the baseline safety requirements that each agent must satisfy as a prerequisite for market entry. The following topics, such as alignment and adversarial robustness, are established areas of AI safety research and a comprehensive review is therefore beyond the scope of this paper. Instead, we briefly define these properties and highlight their specific importance within the proposed multi-agent market context, where interactions are not with a trusted user but with other, potentially untrusted, autonomous agents. These properties form the second layer of the defence-in-depth model, ensuring that the market's components are individually controllable and reliable.

\subsubsection{Adversarial Robustness}

Individual agents must meet minimum standards for robustness against adversarial inputs and sudden environmental changes~\citep{wu2024dissecting}. AI agents may individually implement circuit breakers to prevent harmful actions, in addition to similar market-based measures~\citep{zou2024improvingalignmentrobustnesscircuit}. There should be common standards for establishing these properties, and there should be formally verifiable certificates that are issued to agents that meet the requirements. Agents should be periodically re-certified as the benchmarking capabilities improve and expand.

\subsubsection{Interruptibility}

There should be standardised mechanisms enabling trusted and authorised overseers to safely stop both individual agent actions, as well as the overall agent execution~\citep{orseau2016safely}. It may be necessary to grant some such permissions to automated safety oversight systems as well, that would be able to react at super-human speeds. In terms of collective AI intelligence, interruptibility of individual agents needs to be coupled with systems meant to enable interruptibility of distributed computation, by reliably tracking all of the sub-processes that may have been initiated or delegated to other agents elsewhere within the sandbox. Interruptibility mechanisms ought to be complemented by safe resumption procedures, that would enable the activity to be robustly resumed, if the analysis concludes that the interruption was unwarranted.

\subsubsection{Containment}

Each participating agent should be operating within its own sandbox, in addition to the broader sandbox of the agentic economy. These local sandboxes should enforce strict controls on the behaviour of individual agents, and only permit interactions with the broader market once local safety checks have been satisfied. Such mechanisms would be imperative for localised containment of dangerous or misaligned capabilities and behaviours~\citep{babcock2019guidelines}, though ensuring strict containment requires solving a number of technical challenges, and may not be fully realisable at an individual agent level~\citep{alfonseca2021superintelligence}.

\subsubsection{Alignment}

While the broader market incentive structure aims to mitigate collective misalignment risks, individual agents and components of the ecosystem must all be individually aligned \citep{ji2023ai}. Modern alignment of large models is anchored by preference-based training, starting with supervised fine-tuning followed by optimising a policy against a reward model learned from human preferences, using methods like RLHF \citep{christiano2017deep, ziegler2019fine, stiennon2020learning, ouyang2022training} or direct preference optimisation \citep{rafailov2023direct}. To reduce human load, recent approaches leverage AI feedback constrained by explicit constitutions (Constitutional AI, RLAIF) \citep{bai2022constitutional, lee2023rlaif}. Furthermore, alignment now targets intermediate reasoning through process supervision and step-level verification \citep{lightman2024let}, rather than solely focusing on final outcomes. While prior work has mainly focused on general value alignment \citep{askell2021general} or alignment with specific user intent \citep{goyal2024designing}, individual AI alignment for autonomous task execution within virtual agent sandboxes will likely require further adjustments to these established baselines to address the unique dynamics of inter-agent, rather than human-agent, interaction.

\subsubsection{Mechanistic Interpretability}

Mechanistic interpretability has progressed from reverse-engineering basic feature circuits \citep{olah2020zoom, Elhage2021Framework} to identifying concrete mechanisms like induction heads for in-context learning \citep{Olsson2022InductionHeads}. A key challenge in agent auditing is polysemanticity, where individual neurons encode multiple unrelated concepts due to superposition \citep{Elhage2022Superposition}. Recent capabilities using sparse autoencoders to extract monosemantic features at scale \citep{templeton2024scaling} offer a potential pathway for more precise internal monitoring of agent motivations. Further methods include automated circuit discovery \citep{Conmy2023ACDC}. 

Parallel to these internal methods, monitoring CoT in 'thinking' models presents an opportunity for identifying and correcting misaligned behaviour early in the reasoning process \citep{korbak2025chain, jeung2025safepath}, though long reasoning chains also introduce new surface areas for risk \citep{jiang2025safechain}. However, despite these methodological advances, significant open problems remain regarding the faithfulness and scalability of current interpretability techniques \citep{Rai2024PracticalReviewMI, Sharkey2025OpenProblemsMI}. Therefore, while interpretability methods improve credit assignment and decision auditing, they cannot yet replace comprehensive behavioural benchmarking and must be utilised with an understanding of their current limitations.

\subsubsection{Defence against Malicious Prompts}

Agents must implement robust verification mechanisms to identify attempts to manipulate or jailbreak them during inter-agent interactions. Contemporary defence strategies employ layered approaches, such as using input perturbation to detect adversarial suffixes \citep{Robey2023SmoothLLM}, deploying dedicated 'guardrail' classifiers as a sidecar to the main model \citep{fedorov2024llama}, or utilising separate 'defender' models to gate capability access \citep{Wang2025SelfDefend}. A more robust, structural approach is to enforce safety at the system level, preventing harmful outcomes by design. The CaMeL system~\citep{debenedetti2025defeatingpromptinjectionsdesign}, for example, achieves this by using a dedicated interpreter to enforce information flow control policies, ensuring that even if a model is successfully 'injected', the resulting data flows cannot violate predefined security constraints. Specialised evaluation techniques may also be required to mitigate sophisticated attacks like best-of-N jailbreaking \citep{Armstrong2025DATDP}. Should such attempts be identified, they must be escalated to initiate appropriate measures against the malicious actor.

\subsection{Monitoring and Oversight}

The third layer of the proposed defence-in-depth model focuses on active, real-time detection and response. Preventative measures alone are insufficient to address all potential failure modes, particularly novel emergent behaviours or unanticipated collusion patterns. While the \emph{Market Design} (3.1) section described mechanisms that embed monitoring and auditing into the market's core protocols (e.g., 'Transparency' for logging, 'Reputation' for automated trust assessment, 'Circuit Breakers' for automated halts), this section details the dedicated analytical and governance frameworks that sit above that infrastructure~\citep{busuioc2022ai, holzinger2025human, sharma2024governance, chan2024visibility, kolt2025governingaiagents}. These systems consume the data generated by the market protocols to perform higher-level, active oversight. The following principles (3.3.1-3.3.5) outline this framework, which combines scalable automated monitoring (3.3.1, 3.3.3), proactive adversarial testing (3.3.4), essential post-hoc analytical tooling (3.3.5), and robust, independent human oversight (3.3.2).

\subsubsection{Real-time Monitoring for Systemic Risks}

The agentic sandbox should incorporate mechanisms for continuously tracking interactions, transactions, dependencies, and information flows to identify critical nodes, distribution and utilisation of computational resources, access to shared digital and/or physical goods, concentration of power, and potential pathways through which malicious collusion may occur. Security protocols ought to be established to define and track key risk indicators for networked agentic systems. Bespoke software and AI solutions are needed to facilitate scalable real-time monitoring of high-frequency interactions between agents. This is especially important given that the frequency of these interactions may greatly exceed the capacity of human overseers. Human oversight may therefore need to be complemented and augmented by appropriate tooling implemented at the market level. It may be possible to utilise weaker AI systems to scalably oversee stronger AI systems, though more work is required to fully understand the safety implications of these approaches as the capability of the stronger systems, or AI agent collectives, approaches AGI~\citep{kenton2024scalable, engels2025scaling}. Recursive self-critiquing may improve oversight at higher capability levels~\citep{wen2025scalable}.

Static monitoring targets are susceptible to Goodhart's Law. Advanced agents will likely optimise their behaviour to skirt precisely below static thresholds (e.g., throttling transaction bursts to avoid circuit breakers). Therefore, monitoring regimes must employ dynamic and hidden metrics, akin to financial audits, where the specific trigger conditions for intervention are opaque to the market participants.

\subsubsection{Independent Oversight}

Human oversight of AI agent interactions should be established, and there should be multiple independent~\citep{shneiderman2016dangers, falco2021governing, raji2022outsider} entities with the authority to monitor and investigate multi-agent markets, identify and investigate incidents, and intervene when necessary. These could be certified professionals with sufficient training, equipped with appropriate tools and with clear pathways for escalation. Certified algorithmic oversight may be able to complement the role of human overseers and help scale the volume of monitored agentic interactions. There is also a need to develop mechanisms to protect human operators from manipulation~\citep{carroll2023characterizing, sabour2025human}, social engineering, or information overload by AI agent collectives.

\subsubsection{Detecting Proto-AGI Signatures}

Specific monitoring components ought to be implemented to check for sudden jumps in problem-solving ability across the collective, especially when coupled with resource accumulation and increased coordination. This should include detailed monitoring of the interaction graph to identify sub-graphs~\citep{chen2010dense, yow2023machine} that may be solidifying into an intelligence core, and employing methods that can dynamically detect abrupt behavioural changes~\citep{bridgeford2025detectingperspectiveshiftsmultiagent}. Existing sub-graph extraction methods may need to be extended from their original utilisation in human social networks to properly encapsulate the complexity of agentic interactions in the AI agent sandbox. More emphasis would need to be placed on capability tracking and complementarity, as well as intricacies related to agent identity, as individual agents may get deleted and re-instantiated, copied, or create proxies and sub-agents.

\subsubsection{Red Teaming}

In order to safeguard the market against various threats, it is important to engage in repeated extensive red teaming~\citep{ganguli2022red, ahmad2025openai}. There should be a clear escalation route to ensure that any vulnerabilities that are identified can get patched immediately, and that agent interactions are halted if necessary, to give sufficient time for the updates to be applied, and for recertification of participating agents to be conducted if appropriate. Human red-teaming should be complemented by tiered automated red-teaming via specialised adversarial AI systems, for improved coverage and scalability~\citep{mazeika2024harmbench}. These systems should be expanded beyond identifying the faults of AI agents in isolation~\citep{majumdar2025red}, and perform extensive checks of agentic AI collectives instead~\citep{he2025red}. Red-teaming can also be approached via multi-agent solutions, where a diverse set of AI agents can engage in interactive debate~\citep{asad2025reddebate} to more reliably identify harmful actions and behaviours. To implement these approaches reliably in the agentic sandbox, there needs to be an agreed-upon set of red teaming standards, with associated verifiable certificates for participating agents. This should be complemented by continuous red teaming within the market, aimed at rapidly identifying and mitigating emergent risks.

\subsubsection{Forensic Tooling}

For human overseers to identify root causes of individual failures or systemic cascades, there is a need to develop reliable tooling \citep{jeong2020artificial, alnaqbi2025forensic} for rapid post-incident analysis. This tooling must be capable of parsing large volumes of interaction data to reconstruct causal chains and turn raw traces into legible attack or failure graphs \citep{ezell2025incident, Mukherjee2025ProvSeek}. Standardised logs, capturing not just outputs but also prompts, tool calls, and environmental states \citep{Chernyshev2024LLMForensicsLogs}, would ensure such tooling is broadly applicable across the entire network, enabling analysts to efficiently zoom in on suspicious activity and generate actionable post mortems.

\subsection{Regulatory Mechanisms}

The fourth and final layer of the defence-in-depth model consists of the external regulatory and institutional frameworks that govern the agentic market. While the preceding layers detail the market's internal technical architecture (3.1), component-level requirements (3.2), and active operational oversight (3.3), this layer provides the essential sociotechnical interface with human legal, economic, and geopolitical structures. These mechanisms are not embedded within the market's code but rather enclose it, providing an external source of authority, accountability, and systemic risk management \citep{shavit2023practices, cohen2024regulating, kolt2025lessons}. The following principles (3.4.1-3.4.6) outline the necessary components for legal integration, standardisation, financial remediation of harms, and the mitigation of geopolitical risks.

\subsubsection{Legal Liability and Accountability} 

There should be clear frameworks for assigning liability in case of harm that results from collective actions of agents. In case of distributed and delegated decision-making, no single agent may be fully responsible for the outcome, making auditability, traceability, and explainability a key requirement when permitting consequential actions. Credit assignment~\citep{nguyen2023credit} that aims to associate outcomes with all of the preceding relevant actions is a hard problem even in individual agents, and it would likely be highly non-trivial in the multi-agent setting~\citep{li2025multi}. However, this challenge is not without precedent; legal systems provide a robust model for this in (for example) the form of corporate law, where liability is assigned to the firm - a group agent \citep{list2011group} - as a single legal entity, rather than to its individual employees. This suggests the problem is tractable, requiring the creation of analogous technical and legal structures for agent collectives \citep{list2021group}. In case of patchwork AGI, it would be important to be able to reliably identify all of the responsible agents for each set of actions that correspond to a dangerous capability or a harmful behaviour \citep{franklin2022causal}.

\subsubsection{Standards and Compliance}

There is a pressing requirement for establishing robust standards for agent safety, interoperability, and reporting. These standards must be developed with sufficient foresight to account not only for present-day capabilities but also for rapidly emerging individual agent skills and the potential emergence of patchwork AGI. Beyond mere technical specifications, standards serve as the foundational infrastructure for market-based AI governance, translating abstract technical risks into legible financial risks that can be priced by insurers, investors, and procurers \citep{Tomei2025AIGovernanceMarkets}.

To be effective, these standards ought to be underpinned by rigorous disclosure frameworks that reduce information asymmetry between agent developers and market participants. Such disclosures should cover critical areas including incident reporting, capability evaluations, and resource utilisation, while remaining scale-sensitive to avoid disproportionately burdening smaller innovators \citep{Tomei2025AIGovernanceMarkets}. Enforcement of these standards can then be achieved not just through centralised government action, but through 'regulatory markets', where licensed private entities—such as auditors and insurers—compete to provide oversight and certify compliance with public safety goals \citep{hadfield2023regulatory}.

\subsubsection{Insurance}

Given the difficulties in establishing clear responsibility in collective decision-making scenarios, agentic markets should incorporate insurance mechanisms \citep{lior2021insuring}. Beyond merely providing a compensation pool for harms resulting from misaligned collective behaviours, insurance functions as a critical governance instrument. By setting risk-based premiums and strict underwriting criteria, insurers can incentivise safer development practices and hard-gate market entry for agents that lack sufficient controls \citep{Tomei2025AIGovernanceMarkets}. Possession of appropriate agentic insurance can include policy requirements and premiums scaling dynamically based on the assessed risk level of the agent's certified role and intended tasks \citep{Kvist2025UnderwritingSuperintelligence}.

\subsubsection{Anti-Agent-Monopoly Measures}

A particular risk in the patchwork AGI scenario involves having a group of agents acquire too much power. A patchwork AGI collective could then potentially rapidly outcompete the rest of the market and employ such resources to attempt to resist mitigations in case of harmful and misaligned behaviour \citep{tomei2024machina}. Beyond direct safety risks, uneven access to agentic capabilities may amplify existing socioeconomic inequalities, as entities with superior agent orchestration infrastructure accumulate disproportionate economic advantages~\citep{sharp2026agenticinequality}. Ensuring that no single agent, agent orchestrator, agent-holding entity or agent collective amasses an outsized amount of power and compute within the agentic sandbox, mitigates these concerns. Entities that engage in excessive power-seeking behaviour~\citep{carlsmith2022power, turner2022avoiding} should be flagged for additional checks and inspection. Methods for identifying powerful agents and agent collectives could be built upon prior techniques for identifying power in social networks~\citep{fombrun1983attributions}. Notions of social power, as influence, would need to be complemented by an understanding of the affordances of each agent, and the resources and tools at their disposal. We recommend investigating how established institutional safeguards against centralisation such as levelling mechanisms can be reimagined as protocols to maintain diversity within multi-agent ecosystems.

\subsubsection{International Coordination}

Given the global nature of AGI risks, and the potentially far-reaching consequences of both positive and negative outcomes within distributed agentic markets, international agreements~\citep{scholefield2025internationalagreementsaisafety} and regulatory harmonisation may play an important role in safeguarding against risks \citep{gruetzemacher2023international}. This should also ensure that there are no safe havens for misaligned AI agents or agent collectives, and that all AI agent markets conform to a basic set of safety standards. To ensure compliance with international agreements, verification mechanisms may be required~\citep{wasil2024verification}. For agentic markets and virtual AI sandboxes specifically, there is also a question of localisation - if these virtual entities would span through the international market more freely, or whether they would be contained and regulated within the respective national economies. International coordination around safety may be required in either case, though the details may depend on the exact market model that gets adopted in practice. Thorough harmonisation of standards would enable a potentially more open, interoperable agentic net, without compromising systemic security.

\subsubsection{Infrastructure Governance and Capture}

Implementing this framework requires some degree of centralised enforcement infrastructure. Should agentic economies incorporate too much centralisation, seen as beneficial for effective governance, this would in turn lead to another critical vulnerability: the risk of capture. The integrity of the agentic market depends on the impartial administration of these core components. 

If this infrastructure were to be captured, whether by powerful human interests, or by the emergent patchwork AGI itself, this would also compromise the safety and governance mechanisms, as they may potentially be disabled, bypassed, or in the worst case scenario, weaponised. This highlights a fundamental point of tension between a decentralised vision of the market and the existence of some centralised oversight nodes. Addressing this requires robust socio-technical solutions to ensure that the governors remain accountable and incorruptible.

\section{Conclusion}

The eventual hypothetical development of AGI (or ASI) may not follow the linear and more predictable path of intentionally creating a single, general-purpose entity. AGI, and subsequently ASI, may first emerge as an aggregate property of a more distributed network of diverse and specialised AI agents with access to tools and external models. AI safety and alignment research needs to reflect this possibility, by broadening its scope to increase preparedness for hypothetical multi-agent AGI futures. Deepening our understanding of multi-agent alignment mechanisms is crucial irrespective of whether AGI first emerges as a patchwork, or as a single entity.

The framework introduced in the paper is relevant not only for the emergence of AGI, but also for managing interactions in multi-AGI scenarios (whether interactions are direct or through a proxy web environment and via human users) and, critically, for mitigating the risks of a rapid, distributed transition to ASI via recursive optimisation of the network's components and structure. More specifically, we believe that well-designed and carefully safeguarded market mechanisms offer a promising path forward, and that more AI alignment research should be centred on agent market design, and safe protocols for agent interaction.

Although challenging, this approach offers a potentially scalable path forward. Methodological work on safe market design ought to be complemented by the rapid development of benchmarks, test environments, oversight mechanisms, and regulatory principles that would make these approaches feasible in the future. Many of the measures that we bring up are yet to be fully developed in practice, representing an open research challenge. We would like for this paper to act as a call to action, and help direct the attention of safety researchers towards addressing these challenges and helping design a safe and robust agentic web.

\bibliography{main}

\end{document}